\documentclass[10pt,twocolumn,letterpaper]{article}

\usepackage[pagenumbers]{cvpr} 

\usepackage{graphicx}
\usepackage{caption}
\usepackage{subcaption}

\usepackage{color}
\usepackage{array}

\usepackage{amsmath}
\usepackage{amssymb}
\usepackage{booktabs}

\usepackage{float}

\usepackage[pagebackref,breaklinks,colorlinks]{hyperref}
\usepackage[capitalize]{cleveref}
\crefname{section}{Sec.}{Secs.}
\Crefname{section}{Section}{Sections}
\Crefname{table}{Table}{Tables}
\crefname{table}{Tab.}{Tabs.}

\def\cvprPaperID{10482} 
\def\confName{CVPR}
\def\confYear{2023}

\begin{document}
	
\title{Learning to Kindle the Starlight}

%

%

\author{Yu Yuan$^\dagger$\\
	SJTU\\
	\and 
	Jiaqi Wu$^\dagger$\\
	UESTC\\
	\and 
	Lindong Wang\\
	SJTU\\
	\and 
	Zhongliang Jing$^\ast$\\
	SJTU\\
	\and 
	Henry Leung\\
	UCalgary\\
	\and 
	Shuyuan Zhu\\
	UESTC\\
	\and
	Han Pan\\
	SJTU\\
}

\twocolumn[{%
	\maketitle
	\begin{figure}[H]
		\vspace{-4em}
		\hsize=\textwidth 
		\centering
		\includegraphics[width=15cm]{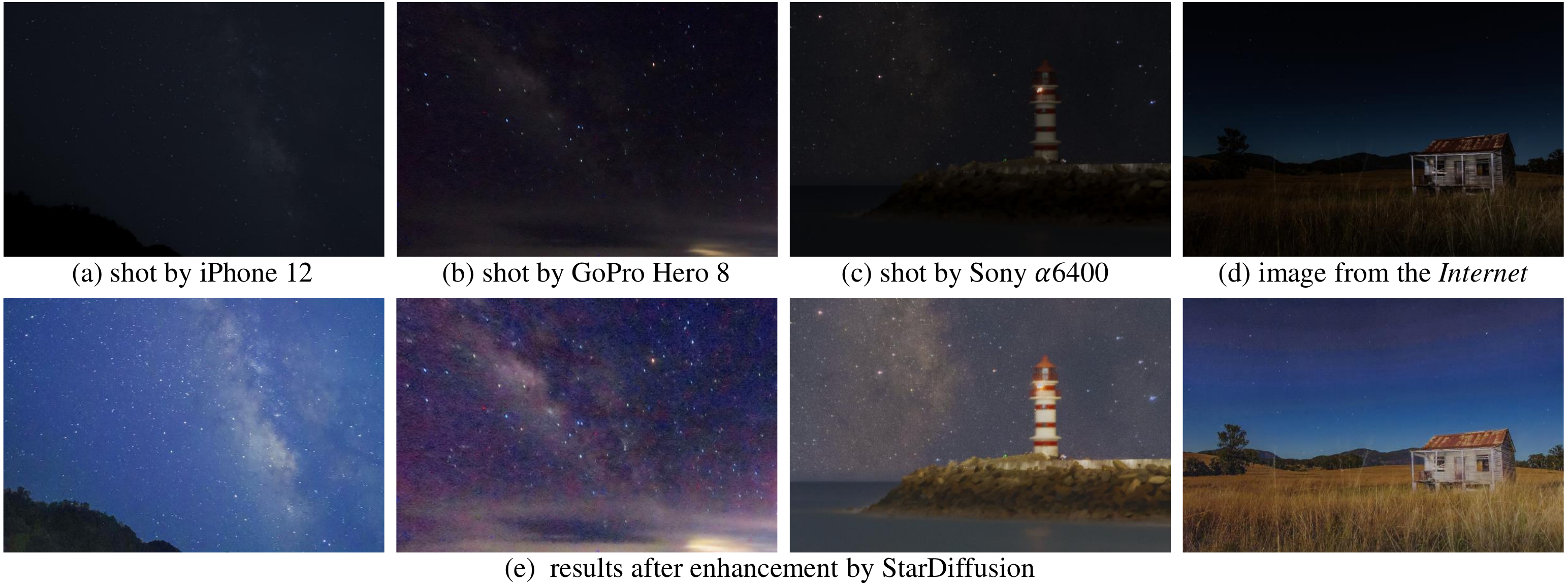}
		\vspace{-1em}
		\caption{Applications of our method. Figure (a), Figure (b), and Figure (c) are the real star field images taken by iPhone12, GoPro Hero 8, and Sony $\alpha$6400, respectively. Figure (d) comes from https://unsplash.com/photos/J8KMIolTmGA. The enhancement results of Figures (a)-(d) by the proposed StarDiffusion are shown in Figure (e).}
		\label{fengmian}
	\end{figure}
}]


\begin{abstract} Capturing highly appreciated star field images is extremely challenging due to light pollution, the requirements of specialized hardware, and the high level of photographic skills needed. Deep learning-based techniques have achieved remarkable results in low-light image enhancement (LLIE) but have not been widely applied to star field image enhancement due to the lack of training data. To address this problem, we construct the first Star Field Image Enhancement Benchmark (SFIEB) that contains 355 real-shot and 854 semi-synthetic star field images, all having the corresponding reference images. Using the presented dataset, we propose the first star field image enhancement approach, namely StarDiffusion, based on conditional denoising diffusion probabilistic models (DDPM). We introduce dynamic stochastic corruptions to the inputs of conditional DDPM to improve the performance and  generalization of the network on our small-scale dataset. Experiments show promising results of our method, which outperforms state-of-the-art low-light image enhancement algorithms. The dataset and codes will be open-sourced.
\end{abstract}

\section{Introduction}
\label{sec:intro}

\begin{figure}[htb]
	\centering
	\includegraphics[width=3.3in]{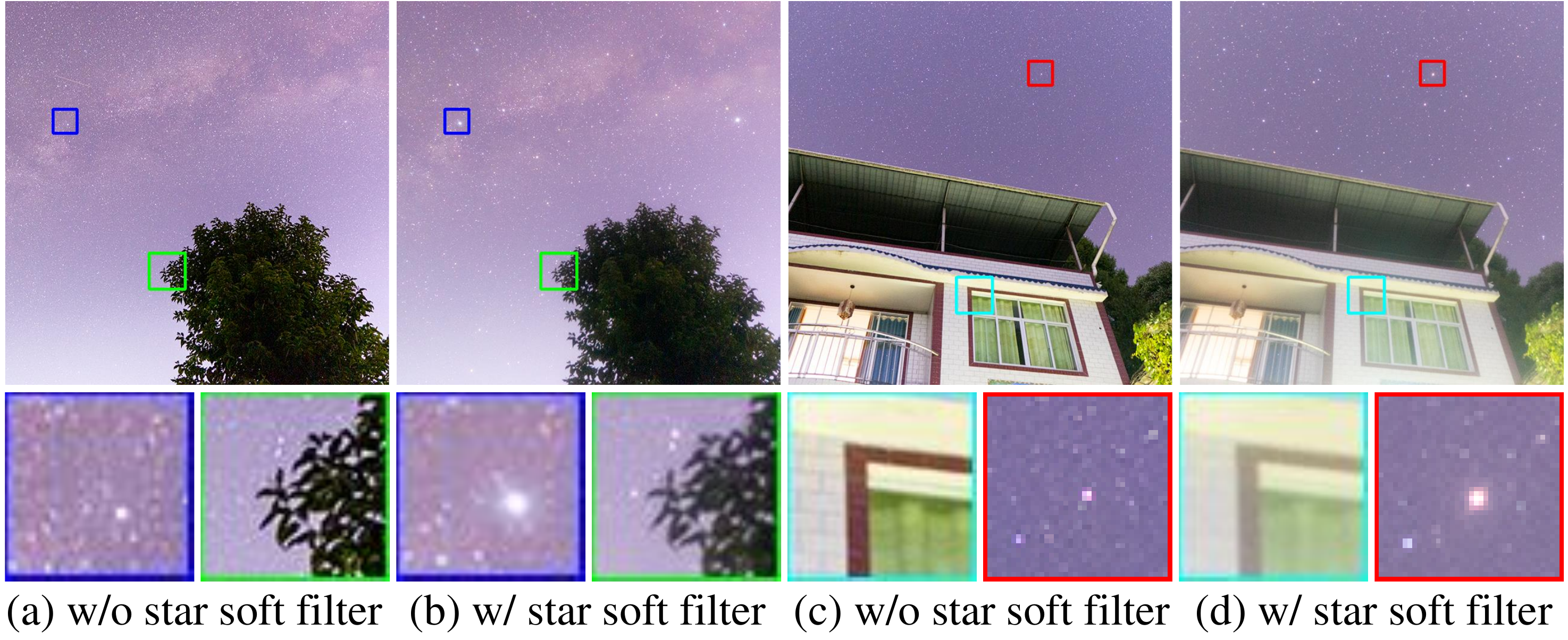}
	\vspace{-1.8em}
	\caption{Using a star soft filter makes the starry sky softer but makes the landscape degraded.}
	\label{roujiao}
\end{figure}

Modern civilization has brought light pollution, which has caused the stars to become very dim. More than 80$\%$ of the world's population and more than 99$\%$ of the U.S. and European populations live under light-polluted skies \cite{science}. For most of us, basking in the sublime glow of the Milky Way has become a luxury, and for star photographers, it's even more of a frustration.\\
\indent As a result, photographers have to set off into untrodden places to capture the star field images with starry sky and commensurate landscapes. They usually utilize expensive large aperture lenses and set the camera gain very high (e.g., ISO 10,000) with long exposure times (e.g., 10 seconds or more) to obtain starlight images with adequate exposure. However, high gains introduce much more noise, and long exposure produces trailing shadows in photographs due to the motions of the stars referring to the ground. Photographers tend to use star soft filters to highlight large stars and make the starlight softer for a better visual effect. However, the use of star soft filters leads to the degradation of image quality as illustrated in  Fig. \ref{roujiao}.\\
\indent Although there are several methods related to star image processing \cite{starnet,tianwen,tiaowu}, there is currently no deep learning method specifically for enhancing star field images due to the lack of training data. The research most relevant to star field image enhancement is low-light image enhancement (LLIE), which aims to improve the perceptual quality of images taken in dark environments \cite{diguang}. According to the network architectures, LLIE approaches can be divided into CNN-based method \cite{llnet, mbllen, lightennet, chen, DRBN, TBEFN, zero, zero+, Retinexnet,kind1, kind+} and GAN-based method \cite{ enlightengan}. Although these methods achieve remarkable success for most low-light scenes, they cannot produce satisfactory results for star field images with both many small stars and huge landscapes (see Fig. \ref{duibi} (b)-(h)).\\
\indent In this paper, we consider the denoising diffusion probabilistic models (DDPM) \cite{DDPM}, which have shown good performance in image generation \cite{DDPM, beat, ILVR,latent,SR3,IDDPM} and image-to-image translation \cite{DDRM, edit,palette,weather,face, repaint}. In conditional DDPM \cite{SR3}, a simple approach is proposed to condition DDPM by concatenating the input with the noisy target image in each reverse process. However, this approach can lead to potential over-fitting, especially for small data sets. To address this challenge, dynamic stochastic corruptions are proposed in our approach. The main contributions of this paper are summarized as follows:
\vspace{-0.6em}
\begin{enumerate}
	\item{We construct the first star field image enhancement benchmark (SFIEB) consisting of 355 real-shot and 854 semi-synthetic image pairs, which makes the comparisons of different LLIE methods on the star field images possible.}
	\vspace{-0.6em}
	\item{We build the first DDPM-based star field image enhancement network, namely StarDiffusion. Specifically, we perform dynamic stochastic corruptions on the inputs of conditional DDPM to improve the learning capability and generalization  of the network on our small-scale dataset.}
    \vspace{-0.6em}
	\item{We conduct comparative experiments with the state-of-the-art LLIE methods. Qualitative and quantitative evaluations verify that the enhanced star field images produced by StarDiffusion achieve the highest perceptual quality. Our method also has good performance for LLIE task.  We also demonstrate the potential of StarDiffusion for enhancing star field photographs taken by consumer-level imaging devices (see Fig. \ref{fengmian}).
		
	}
\end{enumerate}

\section{Related work}

\textbf{Learning-based star image processing} There is seldom work focusing on the processing of star images. Misiura \cite{starnet} proposed a convolution residual net with encoder-decoder architecture to remove stars from nebulae in astrophotography images. For the star image denoising task, Monakhovastar \textit{et al.} \cite{tiaowu} developed a physics-based noise model and used a combination of simulated noisy video clips and real noisy still images to train a video denoiser. Smith \textit{et al.} \cite{tianwen} utilized a diffusion model \cite{DDPM} to generate synthetic galaxy images that are similar to the real data. These star-related works are confined to their respective fields. So far, there is no learning-based method for enhancing star field images.

\textbf{Learning-based low-light image enhancement} Star field image enhancement can be seen as an extreme case of low-light image enhancement (LLIE). LLIE has been widely and intensively studied in recent years \cite{llnet, mbllen, lightennet, chen, DRBN, TBEFN, zero, zero+, Retinexnet,kind1, kind+, enlightengan}. Lore \textit{et al.} \cite{llnet} proposed a deep autoencoder-based method that adaptively brightens low-light images without over-amplifying the brighter parts of the images.  Inspired by Retinex theory \cite{Retinex}, Chen \textit{et al.} \cite{Retinexnet} proposed RetinexNet, which consists of a network for decomposition and a network for  illumination adjustment. Jiang \textit{et al.} \cite{enlightengan} proposed an efficient unsupervised generative adversarial network (GAN) \cite{GAN} that can be trained without paired low-light/normal-light images. Although these methods show satisfactory results for LLIE task, they do not perform well for the star field image enhancement task.

\textbf{Starlight image datasets} Some authoritative datasets already exist in the field of LLIE, such as SID \cite{chen}  and LOL \cite{Retinexnet}. However, most of these datasets consist of indoor scenes, and even with outdoor scenes, there is a lack of images of the night sky.  To the best of our knowledge, the public starlight image training and testing dataset does not exist yet. For this reason, we constructed the first star field image enhancement benchmark (SFIEB).

\section{Dataset collection}
\label{dataset}
\subsection{Real-shot star field image pairs}

\begin{figure}[htb]
	\centering
	\includegraphics[width=3.3in]{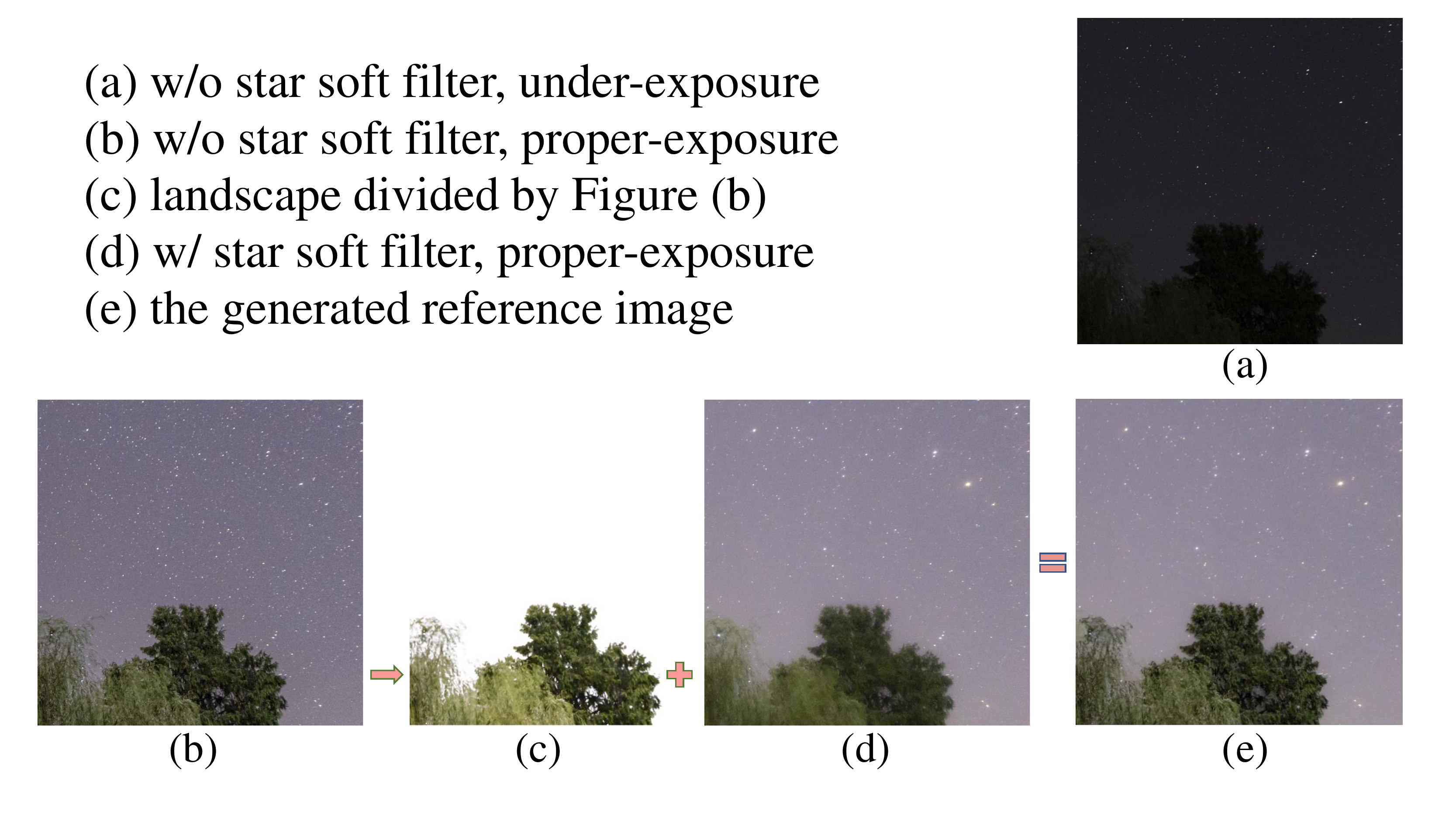}
	\vspace{-1.5em}
	\caption{Process flow of real-shot star field data.}
	\label{zhenshi}
\end{figure}

As shown in Fig. \ref{zhenshi}, we shot three images (Fig. \ref{zhenshi} (a), Fig. \ref{zhenshi} (b), and Fig. \ref{zhenshi} (d)) for a scene in a short duration. More specially, Fig. \ref{zhenshi} (a) taken with under-exposure serves as the input image.
 Fig. \ref{zhenshi} (b) and Fig. \ref{zhenshi} (d) give the pictures which were taken with proper exposure, where the picture of Fig. \ref{zhenshi} (d) was processed by using a star soft filter and generated a more appreciable starry sky while the landscape was degraded. Fig. \ref{zhenshi} (c) shows the landscape divided by Fig. \ref{zhenshi} (b). Using Fig. \ref{zhenshi} (c) to overlay the landscape of Fig. \ref{zhenshi} (d), we obtain the reference image Fig. \ref{zhenshi} (e).

We captured all the images in RAW format with a resolution of
6000×4000 by using a Sony $\alpha$6400 camera. To reduce the possible misalignment in the images, we downsampled all the images to a resolution of 1500×1000. To facilitate training, we further resized and cropped them to patches with a resolution of 640×640 in RGB format. We collected a total of 355 real-shot star field image pairs over a two-year period. However, the diversity of the dataset is still not enough, which needs us to generate more data in a more simple way.

\subsection{Semi-synthetic star field image pairs}
\vspace{-0.2em}
\begin{figure}[htb]
	\centering
	\includegraphics[width=3.3in]{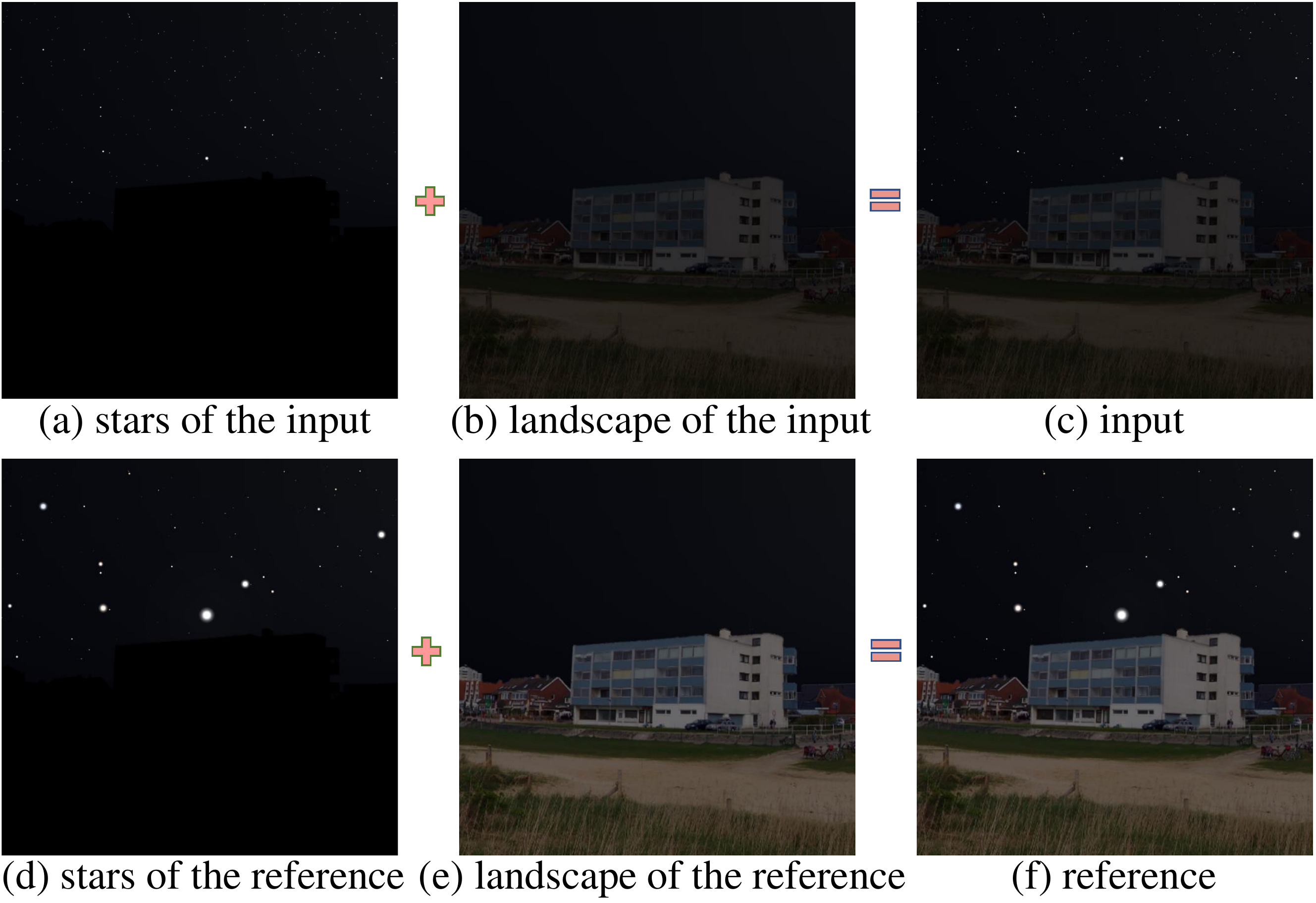}
	\vspace{-1.5em}
	\caption{Process flow of semi-synthetic star field data.}
	\label{fangzhen}
\end{figure}
\begin{figure*}[htb]
	\centering
	\includegraphics[width=6.9in]{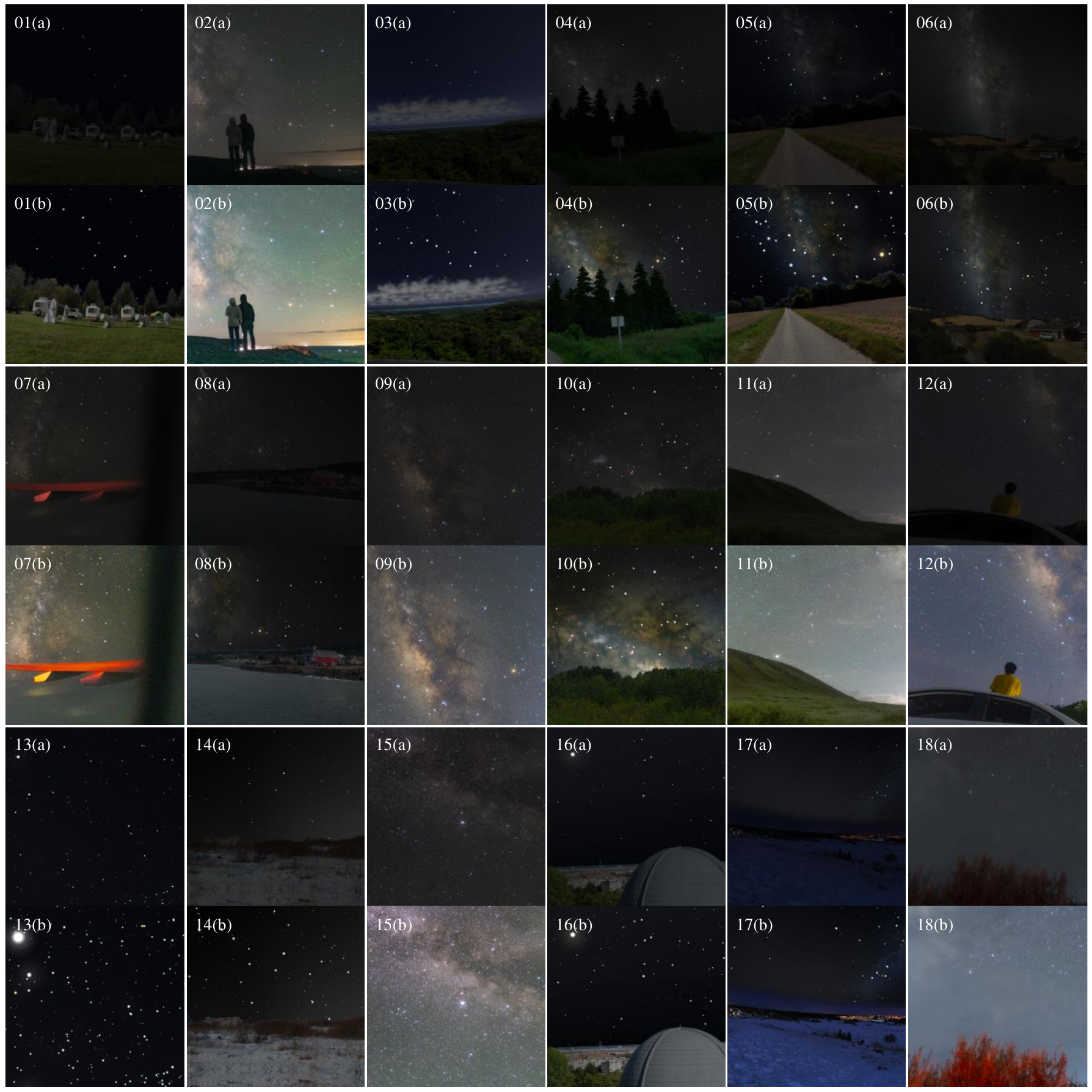}
	\vspace{-1.5em}
	\caption{A montage of sampling image pairs from SFIEB. The rows with the suffix (a) represent the input images, and their next rows with the suffix (b) represent their reference images. The image pairs have been shuffled and the real-shot/semi-synthetic data split is 7/11. A key is provided at the end of this manuscript to indicate which image pairs are real-shot and which are semi-synthetic.}
	\label{shujuji}
\end{figure*}

We leveraged an open-source desktop planetarium software \textit{Stellarium} \cite{stell} to generate synthetic skyscapes of any moment on earth. We collected real landscapes from all over the world from \cite{dijing}. By using the rendering engine of \textit{Stellarium}, we can adjust the brightness and size of the stars and the brightness of the landscapes. As shown in Fig. \ref{fangzhen}, different landscapes and skies are stitched together to generate pairs of input/reference star field images.\\
\indent Following the above way, we obtained 854 semi-synthetic star field image pairs to maximize the diversity of star field. The semi-synthetic data together with the real-shot data constitute the star field image enhancement benchmark (SFIEB), and some samples are given in Fig. \ref{shujuji}.

\section{Background}
\label{sec:format}

\subsection{Denoising diffusion probabilistic models}
\label{ddpmsection}

Denoising diffusion probabilistic models (DDPM) \cite{DDPM, DDPM0} consist of a diffusion process $q$ and a reverse process $p$. The diffusion process is a fixed Markov chain that gradually injects Gaussian noise into a clean image  $\boldsymbol{y}_0$ over $T$ steps, according to a pre-defined variance schedule $\beta_1<\cdots<\beta_T$:
\begin{equation}
	q\left(\boldsymbol{y}_{1: T} \mid \boldsymbol{y}_0\right)=\prod_{t=1}^T q\left(\boldsymbol{y}_t \mid \boldsymbol{y}_{t-1}\right)
\end{equation}
\begin{equation}
	q\left(\boldsymbol{y}_t \mid \boldsymbol{y}_{t-1}\right)=\mathcal{N}\left(\boldsymbol{y}_t ; \sqrt{1-\beta_t} \boldsymbol{y}_{t-1}, \beta_t \boldsymbol{I}\right)
\end{equation}

We can marginalize the diffusion process at each step through:
\begin{equation}
	q\left(\boldsymbol{y}_t \mid \boldsymbol{y}_0\right)=\mathcal{N}\left(\boldsymbol{y}_t ; \sqrt{\bar{\alpha}_t} \boldsymbol{y}_0,\left(1-\bar{\alpha}_t\right) \boldsymbol{I}\right)
\end{equation}
where $\alpha_t=1-\beta_t$ and $\bar{\alpha}_t=\prod_{s=1}^t \alpha_s$.

The reverse process defined by the joint distribution  $p_\theta\left(\boldsymbol{y}_{0: T}\right)$ is also a Markov chain and starts from a standard normal prior $
p\left(\boldsymbol{y}_T\right) $:
\begin{equation}
	p_\theta\left(\boldsymbol{y}_{0: T}\right)=p\left(\boldsymbol{y}_T\right) \prod_{t=1}^T p_\theta\left(\boldsymbol{y}_{t-1} \mid \boldsymbol{y}_t\right)
\end{equation}

\begin{equation}
	p_\theta\left(\boldsymbol{y}_{t-1} \mid \boldsymbol{y}_t\right)=\mathcal{N}\left(\boldsymbol{y}_{t-1} ; \boldsymbol{\mu}_\theta\left(\boldsymbol{y}_t, t\right), \tilde{\beta}_t \boldsymbol{I}\right)
\end{equation}
where $\tilde{\beta}_t=\frac{1-\bar{\alpha}_{t-1}}{1-\bar{\alpha}_t} \beta_t$ and the mean $\boldsymbol{\mu}_\theta\left(\boldsymbol{y}_t, t\right)$ is:
\begin{equation}
	\boldsymbol{\mu}_\theta\left(\boldsymbol{y}_t, t\right)=\frac{1}{\sqrt{\alpha_t}}\left(\boldsymbol{y}_t-\frac{\beta_t}{\sqrt{1-\bar{\alpha}_t}} \epsilon_\theta\left(\boldsymbol{y}_t, t\right)\right)
\end{equation}
where $\epsilon_\theta\left(\boldsymbol{y}_t, t\right)$ is the noise estimated by the neural network. The model is trained by maximizing the variation lower bound of the likelihood $p_\theta\left(\boldsymbol{y}_0\right)$. Similar to \cite{DDPM}, the training target is to minimize :
\begin{equation}\label{loss}
	L(\theta)=\mathbb{E}_{\boldsymbol{y}_0, \epsilon, t}\left\|\epsilon-\epsilon_\theta\left(\sqrt{\bar{\alpha}_t} \boldsymbol{y}_0+\sqrt{1-\bar{\alpha}_t} \epsilon, t\right)\right\|_2^2
\end{equation}

\subsection{Conditional denoising diffusion probabilistic models}
\label{conditionalddpm}
The DDPM is initially proposed for image generation. It needs to introduce conditions to accommodate low-level vision tasks such as image enhancement. Saharia \textit{et al.} \cite{SR3} proposed to implement DDPM by concatenating $\boldsymbol{y}_t$ with input $\boldsymbol{x}$ along the channel dimension in the reverse process $p_\theta\left(\boldsymbol{y}_{t-1} \mid \boldsymbol{y}_t, \boldsymbol{x}\right)$ without modifying the diffusion process:

\begin{equation}\label{88}
	p_\theta\left(\boldsymbol{y}_{t-1} \mid \boldsymbol{y}_t, \boldsymbol{x}\right)=\mathcal{N}\left(\boldsymbol{y}_{t-1} ; \boldsymbol{\mu}_\theta\left(\boldsymbol{y}_t, \boldsymbol{x}, t\right), \tilde{\beta}_t \boldsymbol{I}\right)
\end{equation}
where the mean $\boldsymbol{\mu}_\theta\left(\boldsymbol{y}_t, \boldsymbol{x}, t\right)$ is:
\begin{equation}\label{99}
	\boldsymbol{\mu}_\theta\left(\boldsymbol{y}_t, \boldsymbol{x}, t\right)=\frac{1}{\sqrt{\alpha_t}}\left(\boldsymbol{y}_t-\frac{\beta_t}{\sqrt{1-\bar{\alpha}_t}} \epsilon_\theta\left(\boldsymbol{y}_t, \boldsymbol{x}, t\right)\right)
\end{equation}

\subsection{Proposed method}
\label{methodss}
\begin{figure}[htb]
	\centering
	\includegraphics[width=3.3in]{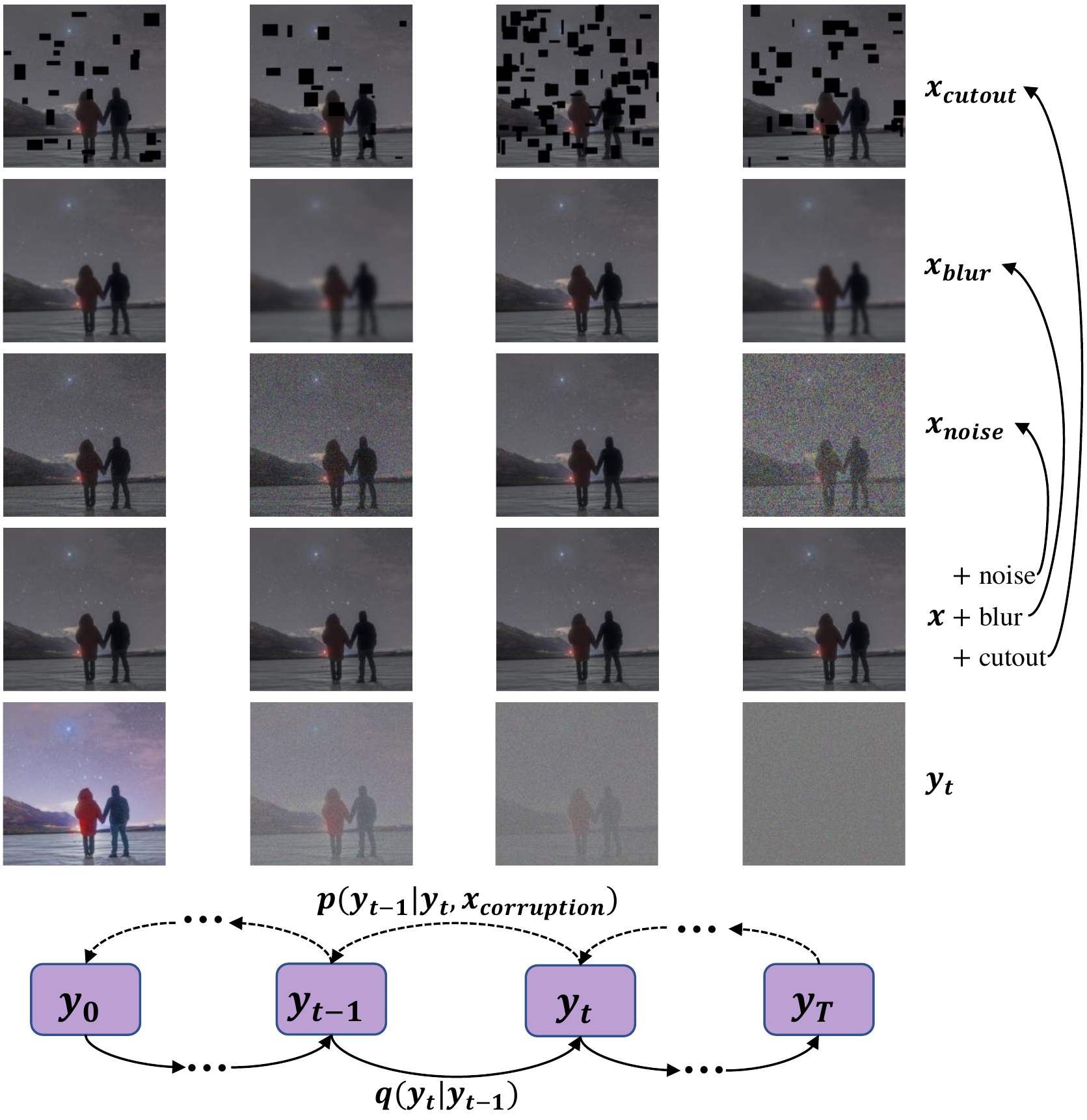}
	\vspace{-1.5em}
	\caption{An overview of the diffusion process (solid line) and reverse process (dashed line) for the proposed conditional DDPM with dynamic stochastic corruptions.}
	\label{tiaojian}
\end{figure}

\begin{figure*}[htb]
	\centering
	\includegraphics[width=6.9in]{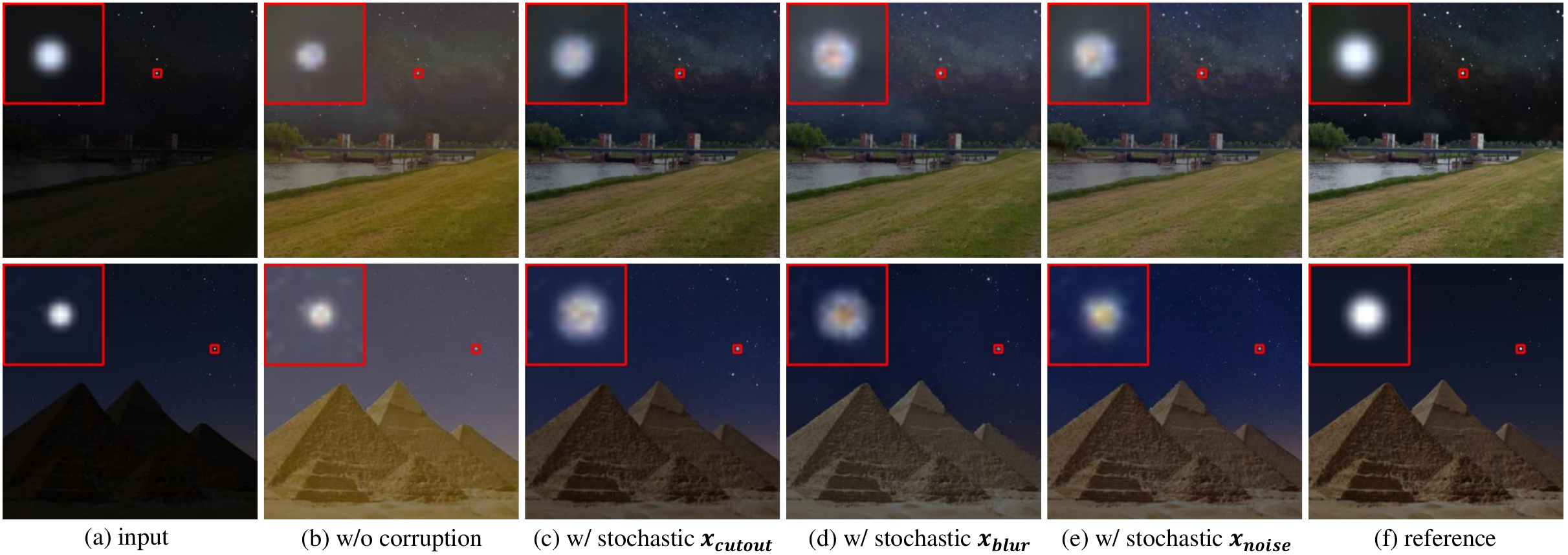}
	\vspace{-1.5em}
	\caption{Qualitative comparisons of different corruption strategies.}
	\label{tiaojian_duibi}
\end{figure*}

\textbf{Conditional DDPM with dynamic stochastic corruptions} The introduction of  constant inputs to the reverse process of conditional DDPM is effective when training on large-scale datasets \cite{SR3}, such as Flickr-Faces-HQ (FFHQ) \cite{FFHQ}, ImageNet 1K \cite{imagenet}, etc. However, after directly adopting this strategy to train on the small-scale SFIEB, we found that the generalization of the network is poor, leading to possible color deviations and no significant increase in the size of the stars in the enhanced images (see the second column of Fig. \ref{tiaojian_duibi}).\\
\indent In the star field images, although the size of stars is small, they are
key visual features due to their high brightness, in contrast to the dark landscapes.
In the conditional DDPM with constant inputs, the network enhances the prominent features (stars) to a lesser extent than the non-prominent features (landscapes).
To improve the network's ability to process stars, we try to weaken the saliency of stars in the inputs.  We implement three forms of stochastic corruptions to the inputs as shown in Fig. \ref{tiaojian}: Gaussian noise,  Gaussian blur, and cutout \cite{cutout}.
 Adding Gaussian noise and applying Gaussian blur can overwhelm star points to some extent, and performing cutout can remove certain regions where stars are located. Although these disruptions are global, they have a greater impact on the stars than the landscapes. Note that the corruptions introduced are dynamic and stochastic, i.e., the corruption added at each step of the reversal process is likely to be different. This dynamic and stochastic strategy can further enhance the diversity and uncertainty of the inputs.  Thus in our approach, the Eq. ( \ref{88} ) and  Eq. ( \ref{99} ) are refined as:

\begin{equation}
	\begin{aligned}
		& p_\theta\left(\boldsymbol{y}_{t-1} \mid \boldsymbol{y}_t, \boldsymbol{x}_{\text {\textit{corruption }}}\right) \\
		=& \mathcal{N}\left(\boldsymbol{y}_{t-1} ; \boldsymbol{\mu}_\theta\left(\boldsymbol{y}_t, \boldsymbol{x}_{\text {\textit{corruption }}}, t\right), \tilde{\beta}_t \boldsymbol{I}\right)
	\end{aligned}
\end{equation}

\begin{equation}
	\begin{aligned}
		& \boldsymbol{\mu}_\theta\left(\boldsymbol{y}_t, \boldsymbol{x}_{\text {\textit{corruption} }}, t\right) \\
		=& \frac{1}{\sqrt{\alpha_t}}\left(\boldsymbol{y}_t-\frac{\beta_t}{\sqrt{1-\bar{\alpha}_t}} \epsilon_\theta\left(\boldsymbol{y}_t, \boldsymbol{x}_{\text {\textit{corruption }}}, t\right)\right)
	\end{aligned}
\end{equation}

Similar to \cite{DDPM}, we use a modified U-Net \cite{unet, DDPM} architecture to implement our method. The network consists of three downsampling blocks, one bottleneck block, and three upsampling blocks. Each downsampling phase consists of two residual blocks \cite{resnet}, a linear self-attention layer \cite{github, attention, linearatt}, and a downsampling operation. The bottleneck consists of a linear self-attention layer sandwiched by two residual blocks. The upsampling phase is mirror-symmetric to the downsampling phase. Our sampling strategy is consistent with that used in \cite{DDPM}. More details are included in the Appendix.

\begin{figure}[htb]
	\centering
	\includegraphics[width=3.3in]{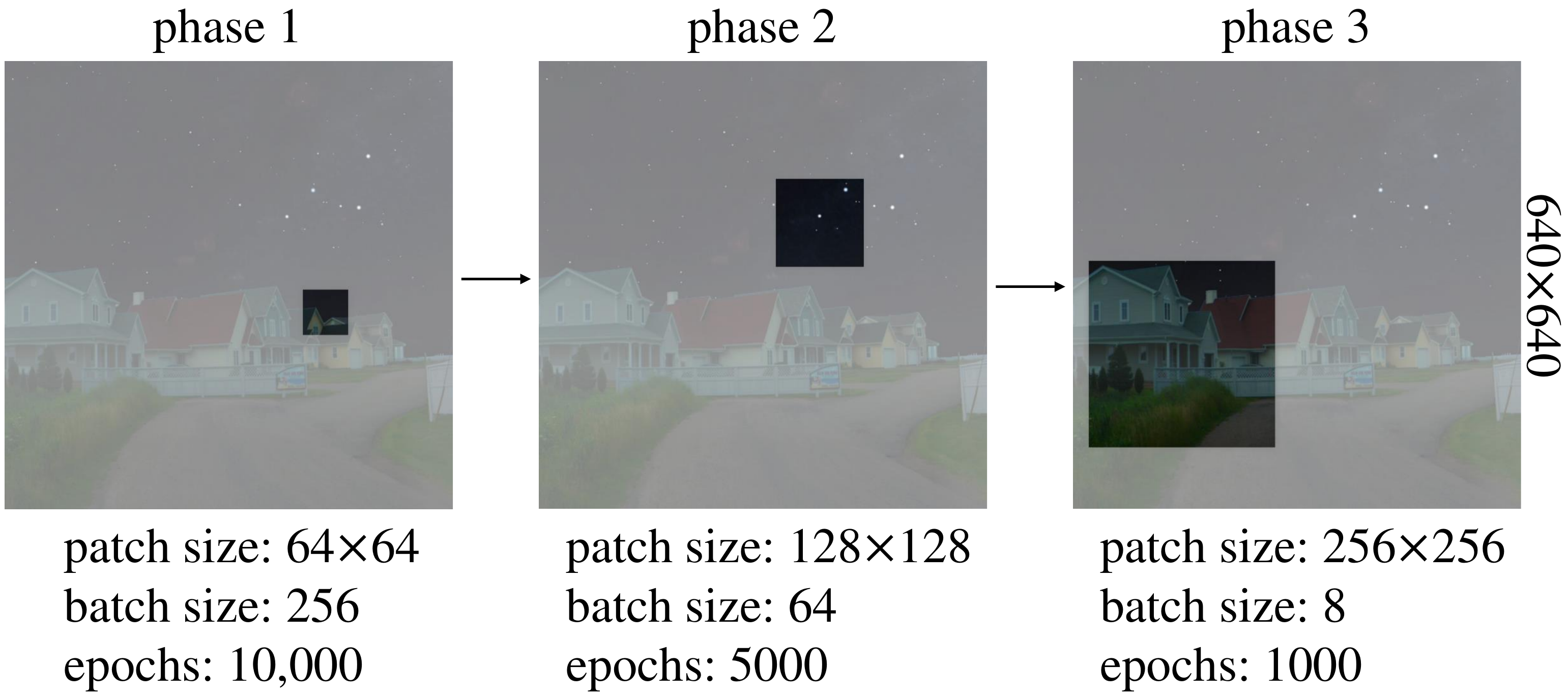}
	\vspace{-1.5em}
	\caption{The proposed cascaded training strategy.}
	\label{cas}
\end{figure}

\textbf{Cascaded training strategy} We proposed a cascaded training strategy shown in Fig. \ref{cas}. It consists of a pipeline of three phases with increasing patch size and decreasing batch size, with each trained model in an earlier phase serving as the pre-trained weight for the next phase. We train for 10k, 5k, and 1k epochs for phase 1, phase 2, and phase 3, respectively. In practice, we find that this strategy speeds up the training process and makes the model more applicable across a range of resolution inputs.

\section{Experiments}
\subsection{Implementation details}

We select 21 image pairs from SFIEB for testing and the rest for training. All the testing image pairs are resized to 512×512 to fit different methods. Our network is trained with an Adam optimizer \cite{adam}. The initial learning rate is set as 1$e$-4 and decreases to 1$e$-6 with the cosine annealing strategy \cite{cosine}. \\
\indent Two metrics including peak signal-to-noise ratio (PSNR) \cite{PSNR} and structural similarity (SSIM) \cite{SSIM} are used to evaluate the enhancement performance of different methods. All experiments are conducted with four NVIDIA Geforce RTX 3090 GPUs and one Intel Core i9-12900k CPU @ 3.70GHz.

\subsection{Corruption details and comparisons}
\label{corruption}

As shown in Fig. \ref{tiaojian}, we set up the following four groups of different corruption strategies: no corruption for the inputs; adding stochastic Gaussian noise with a mean value of 0 and variance range of 10 to 100; applying stochastic Gaussian blur with a kernel size randomly chosen from $\{3, 5,7\}$, and the standard deviations of the corresponding kernels of increasing sizes are 0.8, 1.1, and 1.4, respectively \cite{blur}; performing cutout \cite{cutout, album} in rectangular regions with a number between 1 and 100 while length and width vary between 4 and 32 pixels, and the positions of the cutout are also stochastic. We employ all corruptions with a probability of 0.5.
The introduced dynamic stochastic corruptions match the different patch sizes of inputs at different phases of the cascaded training strategy mentioned in Section \ref{methodss}.  More details can be found in the codes.

\begin{table}[H]
	\caption{Average metric values of conditional DDPM with different corruption strategies for the star field image enhancement task, on 21 sets of testing image pairs. Values in bold indicate the best results.}
	\label{different}
	\centering
	\setlength{\tabcolsep}{2.0mm}{
		\begin{tabular}{l|cc}
			\toprule
			Corruption strategy      	& PSNR $\uparrow$ & SSIM $\uparrow$ \\
			\midrule
			w/o corruption     &  17.9629 &  0.6434   \\
			w/ stochastic	 $\boldsymbol{x}_{cutout}$   &  \textbf{22.7895} & \textbf{0.8073}    \\
			w/	stochastic		$\boldsymbol{x}_{blur}  $     &  20.7954 & 0.7446  \\
			w/	stochastic $\boldsymbol{x}_{noise}  $    &  21.0216 &  0.7955  \\
			
			\bottomrule
	\end{tabular}}
\end{table}

As shown in Table \ref{different}, all introduced corruptions yield higher PSNR and SSIM compared to no corruption. Notably, performing stochastic cutout to the inputs has the best objective performance. We consider that star field images are characterized by heavy spatial redundancy and the network can reconstruct the occluded regions by adjacent pixels. Therefore, the strategy of cutout improves the global representation extraction capability of the encoder and the pixel-level reconstruction capability of the decoder. Fig. \ref{tiaojian_duibi} shows that the introduction of dynamic stochastic corruptions significantly suppresses the color deviations and makes the stars softer and larger. We have also tried mixing different corruptions, but found no significant improvement compared to stochastic cutout, a possible explanation for this is the dominance of stochastic cutout among the three corruption strategies.

\subsection{Comparisons study}

\begin{figure*}[htb]
	\centering
	\includegraphics[width=6.9in]{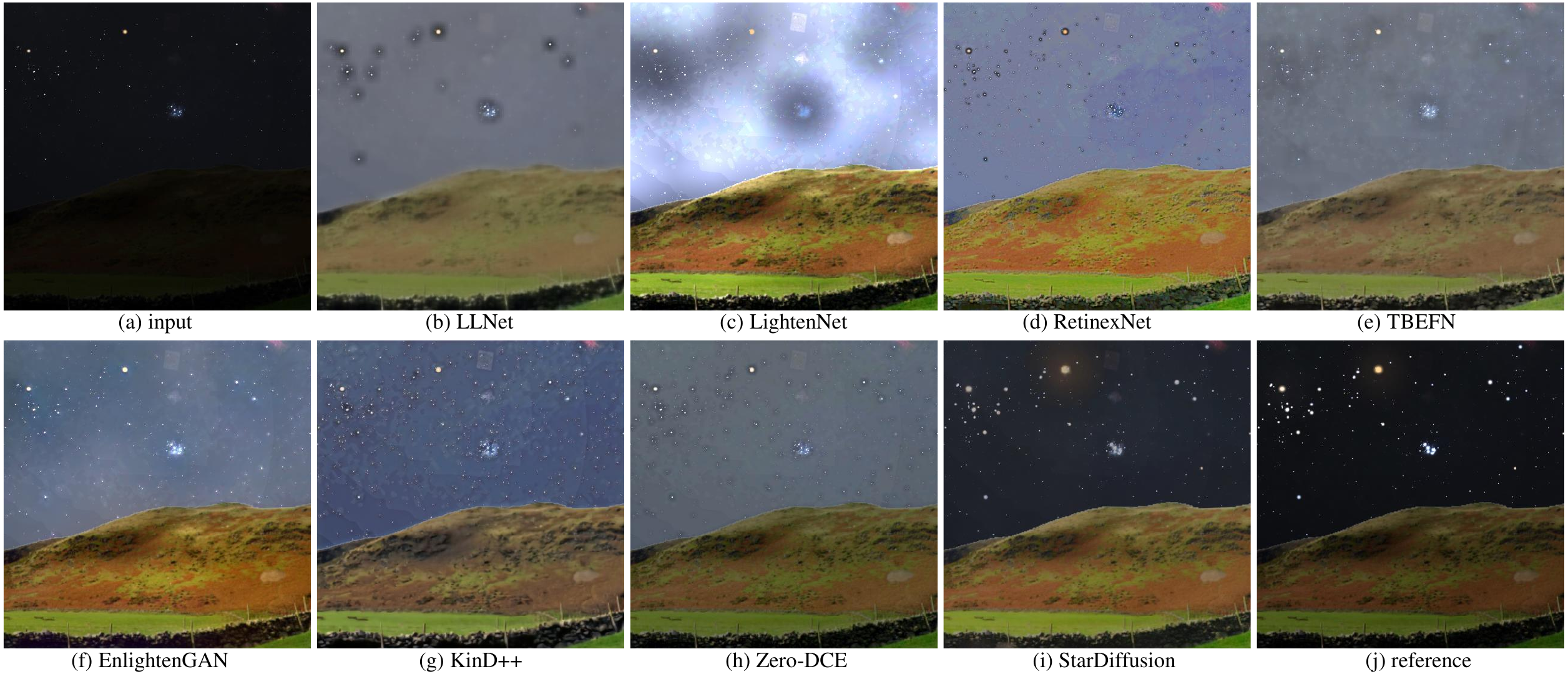}
	\vspace{-2em}
	\caption{Qualitative comparisons between seven LLIE	methods and StarDiffusion on the star field image enhancement task.}
	\label{duibi}
\end{figure*}

\textbf{Comparisons on SFIEB} The comparative study on SFIEB is performed with seven LLIE methods, including LLNet \cite{llnet}, LightenNet \cite{lightennet}, RetinexNet \cite{Retinexnet}, TBEFN \cite{TBEFN}, EnlightenGAN \cite{enlightengan}, KinD++ \cite{kind+}, and Zero-DCE \cite{zero}.  We use StarDiffusion with the corruption strategy of stochastic cutout mentioned in Section \ref{corruption}. The quantitative comparison results on 21 testing images are reported in Table \ref{bijiao}.  Our method is substantially better than other methods in terms of PSNR and SSIM.  As demonstrated by Fig. \ref{duibi}, our results show more natural stars, higher contrast, more continuous colors, and more details.
\begin{table}[H]
	\caption{Average metric values on SFIEB testing set with different methods for the star field image enhancement task.}
	\label{bijiao}
	\centering
	\setlength{\tabcolsep}{1.8mm}{
		\begin{tabular}{r|cc}
			\toprule
			Method          	& PSNR $\uparrow$ & SSIM $\uparrow$  \\
			\midrule
			
			LLNet  \cite{llnet} & 12.5334    & 0.5069   \\
			LightenNet  \cite{lightennet} & 8.7833    & 0.3757   \\
			RetinexNet \cite{Retinexnet}       &  9.4317 &  0.4090  \\
			TBEFN \cite{TBEFN}  &  11.3658 & 0.4959    \\
			EnlightenGAN \cite{enlightengan}    &  11.7933 & 0.5121    \\
			KinD++  \cite{kind+}    &  14.0368 & 0.5433    \\
			Zero-DCE \cite{zero}     &  13.8525 & 0.5641   \\
			
			StarDiffusion    & \textbf{22.7895} & \textbf{0.8073}    \\
			
			\bottomrule
	\end{tabular}}
\end{table}

\begin{figure*}[htb]
	\centering
	\includegraphics[width=6.9in]{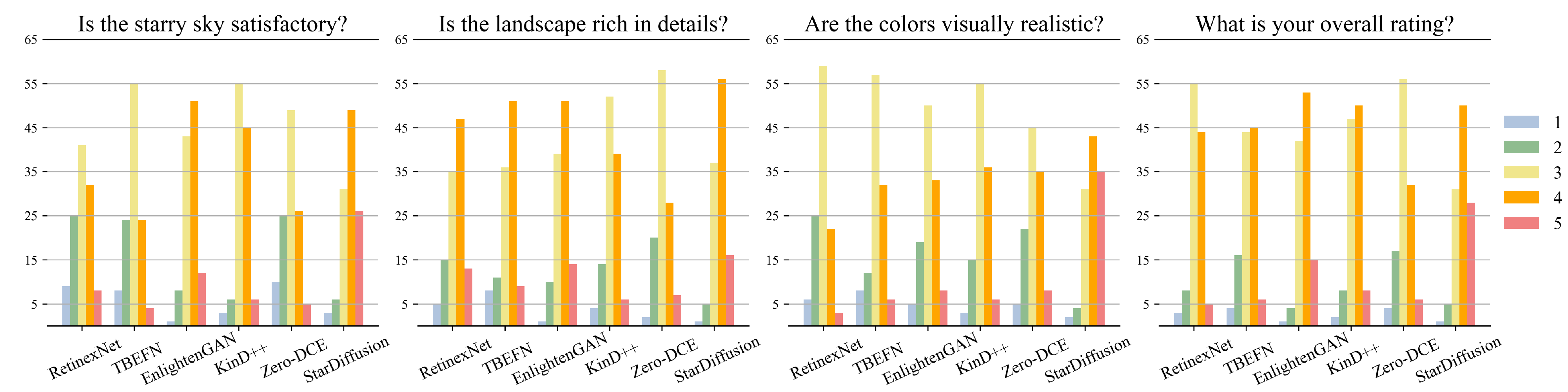}
	\vspace{-2em}
	\caption{Rating distributions for different methods on enhancing star field images in the user study.}
	\label{user}
\end{figure*}
 \begin{figure}[htb]
	\centering
	\includegraphics[width=2.in]{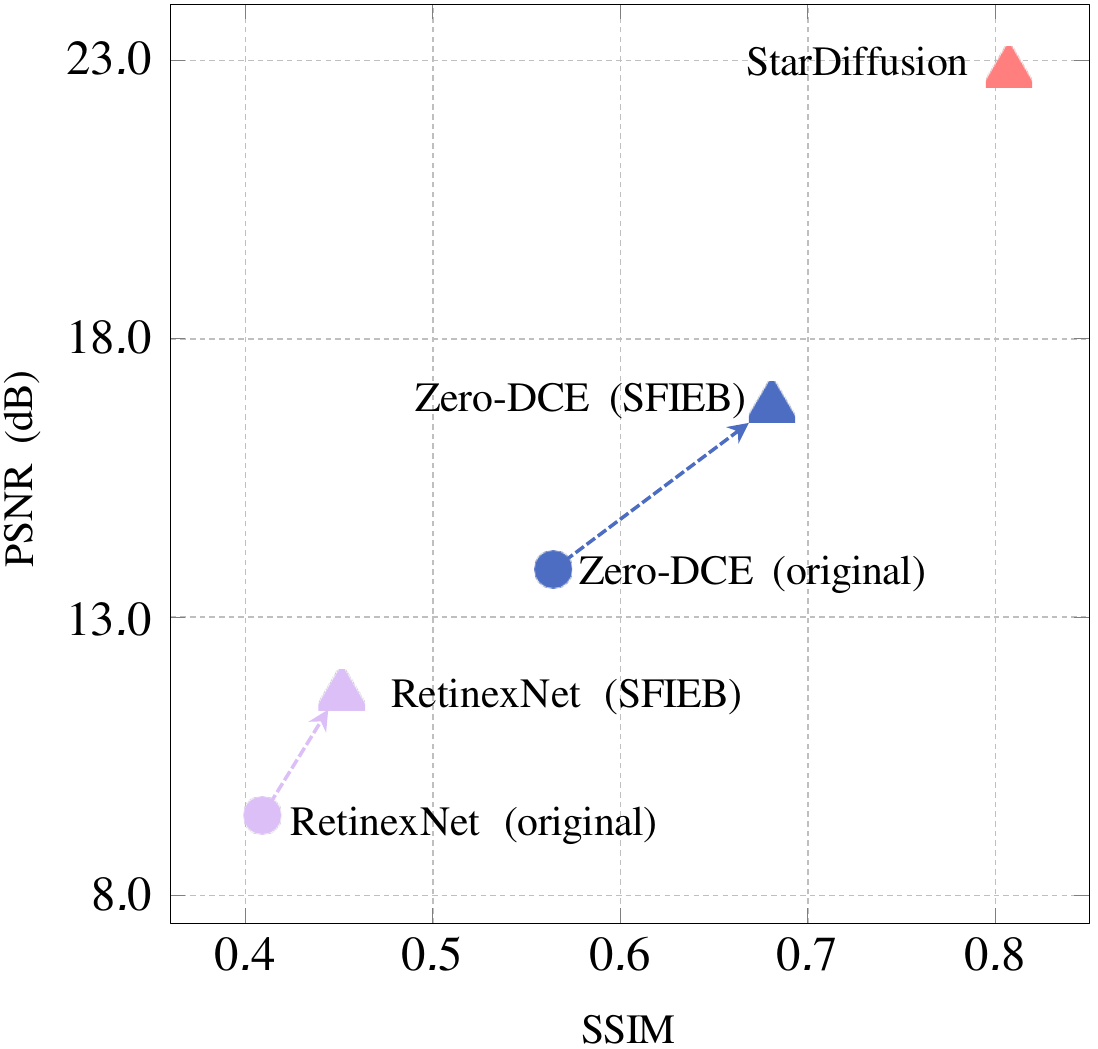}
	\vspace{-0.5em}
	\caption{Effect of two LLIE methods adopting SFIEB. The dots and triangles indicate before and after the SFIEB training set used, respectively.}
	\label{sandian}
\end{figure}
\begin{figure*}[htb]
	\centering
	\includegraphics[width=6.9in]{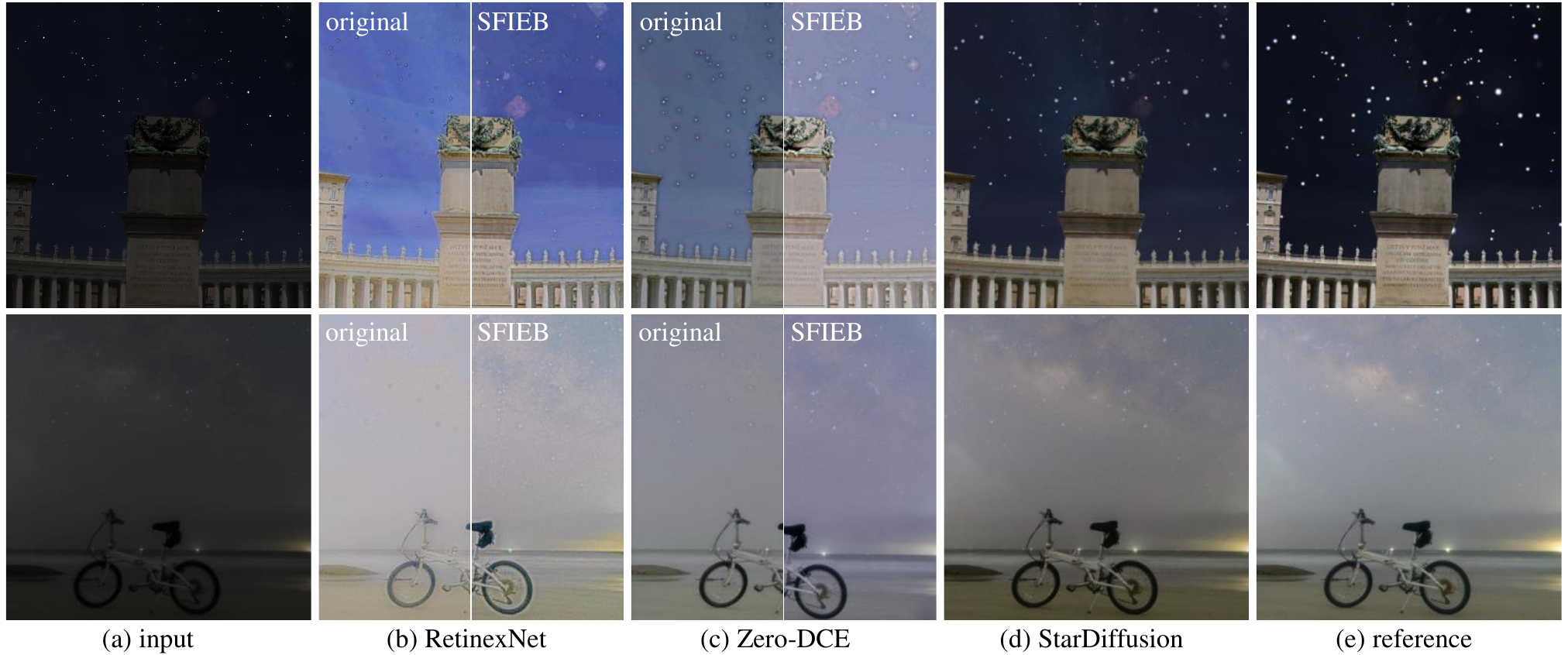}
	\vspace{-2em}
	\caption{Training with SFIEB significantly improves the adaptability of two LLIE methods for the star field image enhancement task.}
	\label{xunlianqianhou}
\end{figure*}

To evaluate the human perception of StarDiffusion and five LLIE methods for enhancing star field images, we conducted a user study with 115 participants in the form of an electronic questionnaire. Fig. \ref{user} illustrates the four questions set in the questionnaire. Ratings are limited to integer score options between 1 (worst) and 5 (best). Overall, StarDiffusion achieves the highest human perception scores, having the most warm colors and the least cold colors.\\
\indent We further trained RetinexNet \cite{Retinexnet} and Zero-DCE \cite{zero} with SFIEB to explore the performance of these two LLIE methods for star image enhancement tasks after training with SFIEB. Fig. \ref{sandian} shows that the two LLIE methods are much better adapted to the star field image enhancement task after sufficient training on SFIEB, but their performance is still inferior to that of StarDiffusion.  Fig. \ref{xunlianqianhou} gives the qualitative improvements of the two LLIE methods after training with SFIEB.\\

\begin{figure*}[htb]
	\centering
	\includegraphics[width=6.9in]{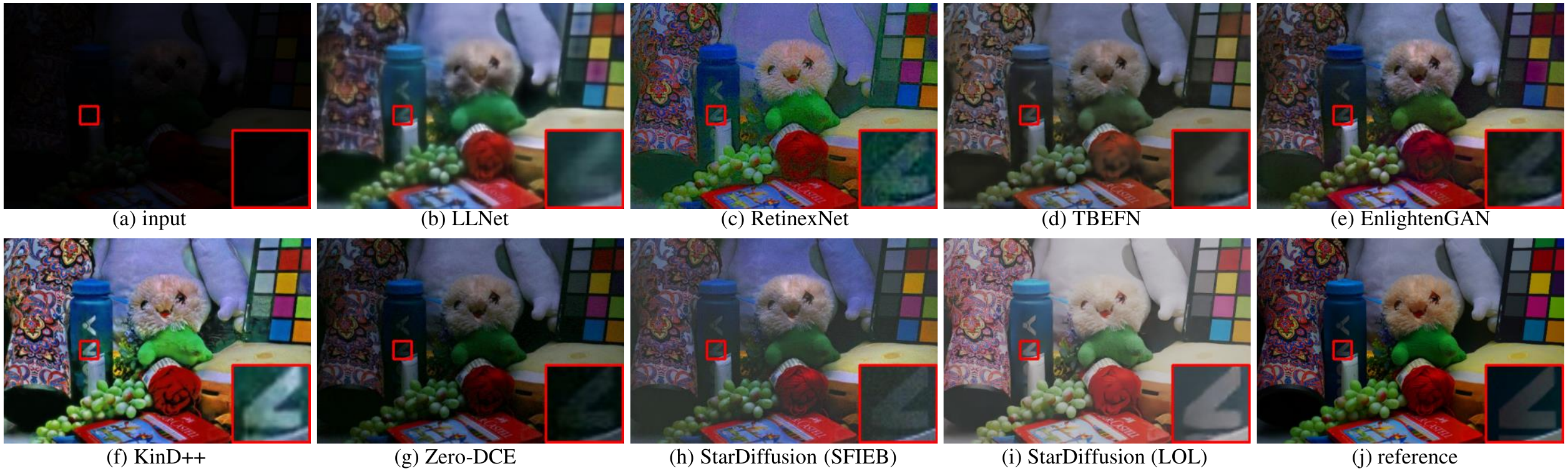}
	\vspace{-2em}
	\caption{Qualitative comparisons on LOL testing set.}
	\label{diguang_lol}
\end{figure*}
\vspace{-1em}
\begin{table}[H]
	\caption{Quantitative comparisons on LOL testing set. Hybrid denotes that EnlightnenGAN uses training data from four datasets \cite{enlightengan} and TBEFN uses training data from two datasets \cite{TBEFN}.}

	\vspace{-0.5em}
	\label{LLIE_LOL}
	\centering
	\setlength{\tabcolsep}{1.0mm}{
		\begin{tabular}{r|c|cc}
			\toprule
			Method          & training dataset & PSNR $\uparrow$ & SSIM $\uparrow$  \\
			\midrule
			LLNet  \cite{llnet} & from \cite{ll_internet} &18.0113    & 0.7258   \\
			RetinexNet \cite{Retinexnet}  &  LOL \cite{Retinexnet}    &  17.6764 &  0.6216  \\
			TBEFN \cite{TBEFN} &   hybrid \cite{TBEFN}  &  17.5638 &  \textbf{0.8001}   \\
			EnlightenGAN \cite{enlightengan} & hybrid \cite{enlightengan}   &  18.1846 & 0.7329   \\
			KinD++  \cite{kind+}  &  LOL \cite{Retinexnet}  &  17.8765 & 0.7536    \\
			Zero-DCE \cite{zero}  & SICE \cite{cai}    &  15.1499 & 0.6883   \\
			
			StarDiffusion & SFIEB   & 18.3263 & 0.6977    \\
			StarDiffusion &  LOL \cite{Retinexnet}  & \textbf{20.7694} & 0.7984  \\
			\bottomrule
	\end{tabular}}
\end{table}
\indent \textbf{Comparisons on LLIE task} StarDiffusion can also be used for LLIE task. We compare StarDiffusion (with stochastic cutout) with six LLIE methods on the LOL \cite{Retinexnet}  testing set. Table \ref{LLIE_LOL} reports that StarDiffusion trained only on SFIEB achieves a competitive PSNR score, and StarDiffusion trained on LOL achieves the highest PSNR score. Fig. \ref{diguang_lol} manifests that StarDiffusion is effective to enhance the lightness of low-light images and reveal more details.
\vspace{-0.4em}
\subsection{Applications}
\vspace{-0.4em}
As shown in Fig. \ref{fengmian}, StarDiffusion effectively improves the visual quality of star field images taken by different consumer-level imaging devices. The proposed SFIEB and StarDiffusion help lower the threshold for capturing high-quality starlight images. More application examples can be found in the Appendix.
\vspace{-0.4em}
\section{Conclusion and discussion}
\vspace{-0.4em}
In this paper, we construct the first star field image enhancement benchmark (SFIEB).
We build the first conditional DDPM-based star field image enhancement network called StarDiffusion. We propose to improve the performance and generalization of the network on small-scale datasets such as SFIEB by performing dynamic stochastic corruptions on the inputs. Experimental results demonstrate that the star field images enhanced by StarDiffusion have better visual quality compared to other LLIE methods. Furthermore, StarDiffusion achieves competitive results on LLIE task. Our method and dataset have potential applications, such as improving the performance of consumer-level devices to capture starry sky scenes. \\
\indent Noise is very common in star field images. However, StarDiffusion does not significantly reduce the extreme noise (see Fig. \ref{fengmian} (e)), and this problem may be caused by the lack of mapping relationships for noise reduction within the image pairs in SFIEB. We will continue to explore this in our future work.\\
\vspace{-0.4em}
\section{Answer key for Fig. \ref{shujuji}}
\vspace{-0.4em}
\label{key}
Real-shot: 02, 07, 09, 11, 12, 15, 18\\
\indent Semi-synthetic: 01, 03, 04, 05, 06, 08, 10, 13, 14, 16, 17


{\small
	\bibliographystyle{ieee_fullname}
	\bibliography{egbib}
}

\end{document}